\documentclass[]{mosi}

\usepackage{array}
\usepackage{tabularx}
\usepackage{enumitem}
\usepackage{float}

\definecolor{oursgray}{gray}{0.95}

\hypersetup{
  colorlinks=true,
  citecolor=mosiblue,
  linkcolor=mosiblue,
  urlcolor=mosiblue,
  pdftitle={Context Scaling: A New Frontier of AI Scaling},
  pdfauthor={Xipeng Qiu},
  pdfsubject={Author-revised preprint of the article published in AI Plus 1 (2026), 100010},
  pdfkeywords={context scaling, AI scaling, context infrastructure, decision-time compute, agents}
}

\title{Context Scaling: A New Frontier of AI Scaling}
\author{Xipeng Qiu\\
  Shanghai Innovation Institute \& Fudan University\\
  \texttt{xpqiu@fudan.edu.cn}\\[6pt]
  {\normalfont\small\parbox{0.88\linewidth}{\centering
  Author-revised preprint. The Version of Record was published in \textit{AI Plus}. This version includes post-publication updates through August 30, 2026.\\[2pt]
  Journal reference: \textit{AI Plus} 1 (2026), 100010. DOI: \href{https://doi.org/10.59717/ipj.aiplus.2026.100010}{10.59717/ipj.aiplus.2026.100010}.}}}

\checkdata[Keywords]{context scaling; AI scaling; context infrastructure; decision-time compute; agents}

\begin{document}

\abstract{%
Many failures of deployed AI arise not from weak inference but from a mismatch between the state a task depends on and the context available when the system acts. Relevant user state, repository revisions, tool traces, sensor readings, or environment changes may be unavailable, distorted, or poorly selected. We call the effort to reduce this bottleneck \emph{context scaling}.

We frame a model as $P_\theta(d\mid c)$ within a system governed by three budgets: training $B_\theta$, decision-time $B_d$, and context-infrastructure $B_c$. Pretraining scaling invests in $\theta$; decision-time scaling searches, refines, or verifies decisions under supplied context; context scaling improves the system's capacity to acquire, represent, deliver, evaluate, and persist deployment state for model use. The three programs interact, but only $B_c$ targets what the system can make available before a decision.

This view makes the model--infrastructure pair the unit of analysis. We define effective-context value as a task- and intervention-dependent attribution, separate deployment state from model-visible context, and identify availability, fidelity, selection, and utilization failures. We then organize current mechanisms through a five-operation context lifecycle and propose controlled budget interventions and utility--cost curves for evaluation. The resulting agenda centers on grounded reporting, interaction-data and full-cost infrastructure, and safety and governance for persistent, adaptive systems.
}

\maketitle

\section{Introduction}
\label{sec:introduction}

Consider one recurring class of failure when a large language model (LLM) is deployed into a real task. A capable base model with a reasonable decision policy can still fail because task-relevant state belongs to the current user, repository revision, tool trace, sensor stream, or environment and is not available in usable form at decision time. We call this the context bottleneck. It is one possible bottleneck alongside model knowledge, reasoning, control, and objective specification; the empirical question is when it becomes the binding one.

The state that deployed systems must reason about is hard to handle: it evolves, has structure, arrives through non-textual channels, and can include tacit or permissioned information that resists direct serialization. Context scaling names the work of reducing this bottleneck by increasing usable decision-time access to deployment state, which takes more than lengthening a text buffer. Its target is higher utility from budgeted context infrastructure.

The dominant accounting of progress emphasizes training investment and, more recently, per-query inference compute. Current external state enters the decision through the infrastructure that constructs model-visible context; the budget for that infrastructure is largely absent from this accounting.

\paragraph{Contributions.}
\begin{enumerate}[leftmargin=*]
  \item We augment the $P_\theta(d\mid c)$ lens with explicit budgets $(B_\theta,B_d,B_c)$, distinguishing training, decision-time, and context-infrastructure investments while retaining $d$ as the output variable.
  \item We define effective-context value as a task-conditional, counterfactual attribution, separate it from raw state and model-visible context, and identify the availability, fidelity, selection, and utilization failures that reduce decision value.
  \item We frame context infrastructure, with agent harnesses as its best-developed runtime form, as the system layer that converts deployment state into model-visible context; four descriptive dimensions identify where different mechanisms are stressed.
  \item We organize the open problems into grounded evaluation and reporting, interaction-data supply and full-cost infrastructure, and safety and governance for persistent and adaptive deployments.
\end{enumerate}

This article is a position paper with a narrative, non-exhaustive survey; it reports no new model or controlled experiment. Context scaling complements pretraining, alignment, and decision-time scaling. Runtime infrastructure remains responsible for acquiring and maintaining deployment-specific state such as the current repository revision, a live sensor reading, or an authorized user history. The survey prioritizes work that isolates a budget in $(B_\theta,B_d,B_c)$, implements a context-lifecycle mechanism, or exposes a measurement, data-supply, safety, or governance constraint; peer-reviewed studies provide the main evidence base, while recent arXiv papers, model reports, and practitioner documentation describe frontier practice.

\paragraph{Testable predictions.} The program predicts positive returns on $B_c$ in controlled comparisons at practically relevant model scales, some transfer of infrastructure improvements within declared task families, and context-value proxies that predict held-out utility better than raw token length or context cost does.

\paragraph{Compact glossary.} We use \emph{deployment state} for the external and persistent state that could inform a task; \emph{model-visible context} for the finite input actually supplied at one invocation; \emph{context infrastructure} for the runtime layer that maps the former into the latter; \emph{context policy} $\pi_c$ for the policy controlling that mapping; \emph{context budget} $B_c$ for the costed capacity spent on acquisition, representation, selection, and persistence; and \emph{effective-context value} for the utility attributed to a context intervention under a declared baseline. Each term names a distinct object.

\section{A Unified View of AI Scaling}
\label{sec:unified-scaling}
\label{sec:formulation}

At the level of one model invocation, an AI model can be described as a conditional distribution over decisions given a model-visible context,
\begin{equation}\label{eq:model-formulation}
  P_\theta(d\mid c),
\end{equation}
where $\theta$ denotes the model's parameters, $c$ the conditioning context (a prompt, prior observations, retrieved documents, a tool-call history, an image, an audio stream, \ldots), and $d$ the decision: a token, a reasoning chain, an action in an environment, or another output the system commits to. Training compute, search, retrieval, and verification enter through the surrounding system, which induces the decision distribution
\begin{equation}\label{eq:system-formulation}
  d \sim Q_{\theta,\pi}(d\mid x; B_\theta,B_d,B_c),
\end{equation}
where $x$ is deployment state, $\pi$ denotes the policies and interfaces surrounding the model, $B_\theta$ is the training investment embodied in $\theta$, $B_d$ is per-decision compute for proposing, searching, and verifying outputs, and $B_c$ is the budget for acquiring, representing, selecting, and persisting model-visible context. The three budgets can interact, and many systems jointly vary more than one.

The system-level distribution still leaves its two context-side objects informal: the deployment state $x$ that a system can draw on, and the value created when part of that state reaches the model as usable context. Defining both makes the return on $B_c$ measurable before the three investment programs are compared.

\subsection{From World State to Effective Context}
\label{sec:context-definition}
\label{sec:context-environment}

Context should be distinguished from environment. The environment is the external task world in which the system acts: a repository and terminal for a coding agent, a physical scene for a robot, or a user's evolving work state for an assistant. We treat deployment as a sequence of decision steps indexed by $t$; at each step the model commits a decision $d_t$ conditioned on model-visible context $c_t$, produced from deployment state by the surrounding infrastructure. The environment is larger than this representation, dynamic, and only partially observed. The distinction between $c$ and $d$ is local to a model invocation: across a trajectory, prior decisions, tool results, and execution traces become part of later context.

\paragraph{Context infrastructure.}
The model-level $c_t$ is the output of context infrastructure: the system layer that decides what state reaches the model at decision time. Increasing $B_c$ expands the budgeted capacity of this mapping.

Model-visible context varies in utility. Four failures can intervene between deployment state and task utility. \emph{Availability failure} leaves relevant state unreachable; \emph{fidelity failure} distorts it during encoding or summarization; \emph{selection failure} excludes, delays, or poorly places represented state under the read budget; and \emph{utilization failure} occurs when the model fails to use faithfully delivered evidence. The first three are primarily infrastructure-side failures, while utilization depends strongly on the model; in practice the boundary is joint because selection and presentation can change what a fixed model uses.

\paragraph{Deployment state, model-visible context, and effective-context value.}
Three objects should be separated. Let $x_{\mathrm{raw}}$ denote the deployment state and affordances that could in principle inform a decision: retrieved passages, tool outputs, observations, memory entries, logs, interaction histories, tool schemas, and handler states. A context-production mapping $\mathcal{I}_{\pi_c}$ acquires and represents parts of this heterogeneous state, then selects a finite model-visible input $c_{\mathrm{window}}$ under budget $B_c$:
\[
  c_{\mathrm{window}} = \mathcal{I}_{\pi_c}(x_{\mathrm{raw}};B_c).
\]
The mapping transforms deployment state: a sensor stream may become an embedding, a tool trace may become a structured record, and several sources may become one summary. The model conditions on $c_{\mathrm{window}}$. We reserve \emph{effective-context value} for the utility attributed to that input after fixing a task distribution, model, decision procedure, utility function, and intervention baseline. Write $U(\tau)$ for trajectory utility and $b$ for a declared baseline context. The effective-context value of an intervention producing $c$ is
\[
  V_{\mathrm{eff}}(c;b) = \mathbb{E}[U(\tau)\mid c] - \mathbb{E}[U(\tau)\mid b].
\]
Here and below, the conditioning notation is shorthand for outcomes under the declared controlled context replacement, not observational conditioning on naturally occurring context. It is therefore a task-conditional causal estimand defined over interventions on the model-visible input. When a declared unit $i$ (a passage, source, memory slot, sensor channel, or structured field) is ablated or replaced, its marginal attribution is
\[
  \Delta U_i \;=\; \mathbb{E}\big[\,U(\tau)\mid c\,\big] \;-\; \mathbb{E}\big[\,U(\tau)\mid c \setminus \{i\}\,\big],
\]
where $c\setminus\{i\}$ denotes the pre-specified replacement intervention, which may replace a source or representation instead of deleting tokens.

\begin{figure}[ht]
\centering
% Vector figure (default font, pdflatex build). Editable source: 14_latex_template/figures-src/fig-effective-context.tex
\includegraphics[width=\linewidth]{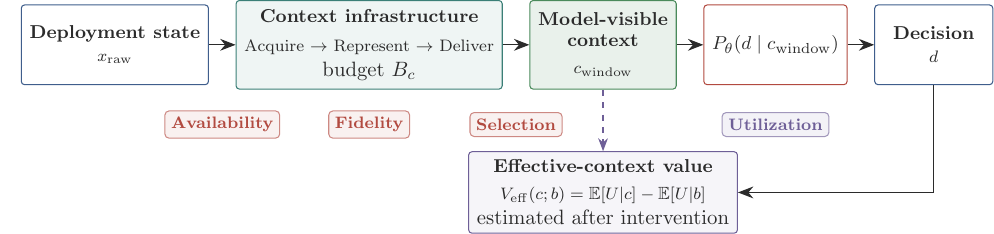}
\caption{Deployment state $x_{\mathrm{raw}}$ is transformed by context infrastructure into model-visible context $c_{\mathrm{window}}$, on which $P_\theta$ actually conditions. Effective-context value $V_{\mathrm{eff}}$ is estimated afterward from utility under declared interventions and baselines. Availability, fidelity, and selection failures occur before the model invocation; utilization failure occurs when faithfully delivered evidence fails to improve the model's decision.}
\label{fig:raw-effective-context}
\end{figure}

Synergy can be represented by averaging marginal contributions over entry orders, following a Shapley-style allocation \citep{shapley1953value,lundberg2017shap}; the result still depends on the chosen units, replacement baseline, and sampled coalition distribution. Interventions on latent, compressed, or multimodal state should target a source, channel, memory slot, or representation directly. Negative contributions and redundancy remain part of the estimate. Figure~\ref{fig:raw-effective-context} summarizes the transformation and the four failure points.

\section{Three Scaling Programs}
\label{sec:three-programs}

\begin{figure}[ht]
\centering
% Vector figure (default font, pdflatex build). Editable source: 14_latex_template/figures-src/fig-regimes.tex
\includegraphics[width=\linewidth]{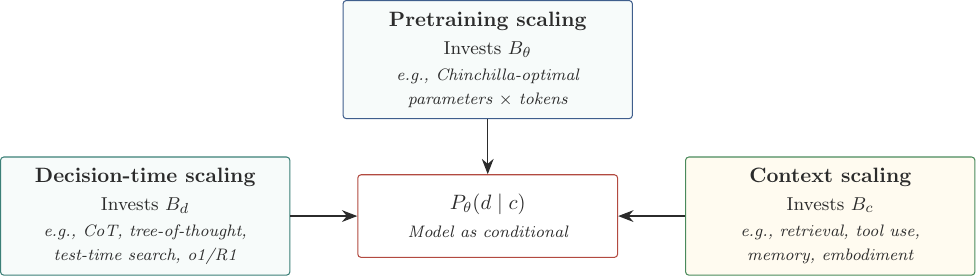}
\caption{Three coupled investment programs around a model invocation $P_\theta(d\mid c)$. Pretraining spends $B_\theta$ on the base model; decision-time scaling spends $B_d$ on proposing, searching, or verifying decisions under a protocol-fixed context; context scaling spends $B_c$ on the infrastructure that maps deployment state into model-visible context. The budgets are reported separately because their returns can interact.}
\label{fig:three-regimes}
\end{figure}

The three-budget lens sorts recent progress into three interacting investment programs. Pretraining scaling spends $B_\theta$ to improve the base conditional over a wide range of $(c,d)$ pairs. Decision-time scaling spends $B_d$ to explore, refine, or verify candidate decisions under model-visible context held fixed by protocol. Context scaling spends $B_c$ to change the mapping from deployment state $x$ to model-visible context $c$. Figure~\ref{fig:three-regimes} places the three programs around the same model invocation.

\paragraph{Demarcation by counterfactual budget.} For an intervention that touches several components, classification is secondary to attribution. We first report which of $(B_\theta,B_d,B_c)$ changed and estimate a response surface $U(B_\theta,B_d,B_c)$ on a declared task family. A single-axis label is used only when two budgets are held fixed by design and the third is the manipulated resource.

\begin{tcolorbox}[colback=blue!4, colframe=black!55!white, arc=2mm, boxrule=0.5pt, breakable, title={\bfseries Demarcation rule}]
To attribute a gain, hold two budgets fixed and vary the third. Report mixed interventions by the budgets they change:
\begin{itemize}[leftmargin=*, nosep]
  \item vary training data, parameters, or optimization while holding the runtime protocol fixed $\Rightarrow$ estimate the marginal return on $B_\theta$;
  \item vary per-decision search, sampling, rollout, or verification while holding model-visible context fixed $\Rightarrow$ estimate the marginal return on $B_d$;
  \item vary acquisition, representation, selection, or persistence capacity while holding the model and decision procedure fixed $\Rightarrow$ estimate the marginal return on $B_c$.
\end{itemize}
\end{tcolorbox}

Take the two cases that most often blur the boundary. Reasoning reinforcement learning (RL; o1- and R1-style \citep{openai2024o1,deepseek2025r1}) changes $B_\theta$ during training and often enables a larger effective $B_d$ at deployment; its total gain should therefore be decomposed with both budgets reported. Tool-use and retrieval-policy RL (e.g.\ Search-R1 \citep{jin2025searchr1}) changes $B_\theta$, uses $B_d$ to choose queries or calls, and consumes $B_c$ to return evidence. Withholding evidence establishes the value of external state; budget attribution additionally requires varying retrieval capacity or evidence quality at matched model and decision budgets and reporting interaction effects.

\subsection{\texorpdfstring{Pretraining Scaling: Scaling $\theta$}{Pretraining Scaling: Scaling theta}}
\label{sec:pretraining}

Empirical scaling laws describe how pretraining loss decays as a joint function of parameters and tokens under a fixed compute budget \citep{kaplan2020scaling,hoffmann2022chinchilla}; much generic model-improvement investment has therefore gone into larger models, datasets, and training runs. Data-constrained scaling work suggests that high-quality natural data is becoming a tighter resource, while filtering, synthetic augmentation, and multi-epoch training alter that constraint \citep{muennighoff2023dataconstrained,sorscher2022pruning}. Situated deployments also require current interaction history, tool feedback, user-authorized state, or post-cutoff information beyond the pretrained prior.

\subsection{\texorpdfstring{Decision-Time Scaling: Inference Compute on $d$}{Decision-Time Scaling: Inference Compute on d}}
\label{sec:decision-time}

Decision-time scaling varies the inference compute spent per decision under a protocol that holds the model and supplied external evidence fixed. Chain-of-thought prompting unrolls an intermediate reasoning trajectory \citep{wei2022cot}; self-consistency, tree-of-thought search, and verifier- or process-reward-guided decoding sample, search over, and score multiple trajectories \citep{wang2023selfconsistency,yao2023tot,lightman2023prm,cobbe2021verifier}; reasoning-specialized models can allocate more test-time computation per query \citep{openai2024o1,deepseek2025r1}, with empirical compute--performance curves in restricted settings \citep{snell2024testtimecompute}. Current repository changes, sensor readings, and user actions enter the decision only through acquisition, which spends $B_c$ and often $B_d$ on information-seeking actions.

\subsection{\texorpdfstring{Context Scaling: Investing in $B_c$}{Context Scaling: Investing in Bc}}
\label{sec:context-scaling}

The third program, the subject of this paper, invests the infrastructure budget $B_c$ in constructing the conditioning variable $c$.

Context scaling covers budgeted investments that change the information state on which decisions are conditioned. Its scope includes long-context mechanisms and extends to retrieval, memory, tools, and grounded observation; a million-token window filled with static documentation can increase token count while adding little task utility. Agentic decision-making has a broader scope that also includes goals, planning, and action selection.

We therefore define context scaling as a budgeted intervention on the capacity to acquire, represent, select, or persist deployment state for model use.

\subsubsection{Measuring and Optimizing Context Value}

The counterfactual definition yields intervention-specific estimates of context value. \emph{Declare} the task distribution, utility horizon, model checkpoint, context units, replacement baseline, evaluator, and all three budgets. \emph{Intervene} with a randomized or matched design on one context-infrastructure capacity at several $B_c$ levels; when the intervention also changes tool calls, rollout length, or verifier use, report the corresponding change in $B_d$ and estimate the interaction. \emph{Measure} a utility--cost curve and uncertainty intervals, using grouped or randomized subset interventions when exact Shapley attribution is infeasible. \emph{Report} the marginal utility per unit context cost and, optionally, a decision-relevance density (the fraction of declared units whose specified replacement lowers utility beyond a stated tolerance). Density is comparable within protocols that share the same units, tokenization or representation, replacement baseline, task distribution, and tolerance; compression, duplication, or re-partitioning can change density while preserving information. Cross-domain comparison should therefore use disclosed cost and utility curves.

The context policy $\pi_c$ chooses infrastructure actions to maximize expected task utility minus deployment-specific context costs. Retrieval, tool-use, memory-write, active-sensing, and context-budget policies expose different learning signals \citep{jin2025searchr1,wei2025webagentr1}; sparse terminal rewards can discourage informative questions because they fail to distinguish useful queries along the trajectory \citep{zou2026selflocking}. Objective design also faces a substantive tension between answer accuracy, which rewards resurfacing known content, and information gain, which rewards reaching new material.

\subsubsection{What Kind of Scaling Law Exists?}

Mechanism-specific response curves currently provide the closest counterpart to the Kaplan or Chinchilla scaling laws. $B_c$ purchases heterogeneous capacities, and $c$ is a structured information state. Here, a scaling program means the systematic estimation of marginal utility as $B_c$ varies at matched $B_\theta$ and $B_d$.

\paragraph{Restricted scaling laws.} The context-aware scaling law of \citet{montgomery2025contextaware} models downstream task performance as the product of two saturating power laws (one in training compute, one in prompt context length) times a penalty when the prompt exceeds the model's context window. The form fits in-distribution downstream performance across three orders of magnitude of training compute on three tasks, with a task-dependent characteristic context length. A complementary line \citep{intrinsic2025contextbounds} argues from an intrinsic-entropy framework that training-data volume bounds the optimal context length. Together they characterize restricted context coordinates and their interaction with training scale.

\paragraph{Empirical patterns.} Dense retrieval exhibits regular scaling with model size and annotation budget \citep{fang2024densescaling}. LongPPL predicts long-context benchmark scores more reliably than standard perplexity in the studied setting \citep{longppl2024}. Frontier technical reports document growing context windows and agent-oriented context training \citep{kimi2025k15,deepseek2026v4,glm2025glm45,zai2026glmrepo}, and benchmarks such as RULER and LongBench v2 separate window capacity from evidence utilization \citep{hsieh2024ruler,bai2024longbenchv2}.

\begin{tcolorbox}[colback=blue!4,colframe=black!55!white,boxrule=0.5pt,arc=2mm,breakable,title={\bfseries Observed returns on context investment}]
\small
\begin{itemize}[leftmargin=*,itemsep=3pt,topsep=2pt]
  \item \textbf{Datastore capacity.} \citet{shao2024retrievalscaling} report monotonic gains as a retrieval datastore grows, including settings where a smaller retrieval-augmented model outperforms a larger standalone model at matched training compute. Runtime retrieval and storage costs determine the resulting budget substitution.
  \item \textbf{Learned retrieval policy.} \citet{jin2025searchr1} report 20--41\% relative gains over retrieval-augmented baselines on factual QA. The intervention spans $(B_\theta,B_d,B_c)$.
  \item \textbf{Retrieval and generation allocation.} \citet{yue2025ragscaling} hold the base model fixed and vary a budget distributed across retrieved documents and generation, tracing a coupled $(B_c,B_d)$ frontier.
\end{itemize}
\end{tcolorbox}

\subsubsection{Mixed Budget Interventions}
\label{sec:mixed-cases}

The budget view reports mixed cases by expenditure. Domain-adaptive continued pretraining varies $B_\theta$ under transfer-style laws \citep{hernandez2021transfer}. In-context learning changes model-visible context and may also change generation cost. Retrieval and tool use spend $B_d$ on query or action selection and $B_c$ on interfaces, returned evidence, and state handling. Chain-of-thought primarily varies $B_d$ when no external observation is acquired; a world-model rollout combines decision-time simulation with a learned proxy for environment transitions. Persistent memory spends $B_c$ on storage, validation, and recall, while continual weight adaptation also changes $B_\theta$.

\paragraph{Complementarity and interaction.} Returns on $B_c$ are coupled to both $B_\theta$ and $B_d$: richer context pays when the model and decision procedure can use it, and some context gains should shrink as models improve. The resulting object is a joint response surface whose $B_c$ slope can vary with model scale and decision-time compute. This coupling also yields a falsification condition: if controlled returns on $B_c$ vanish at matched $B_\theta$ and $B_d$, a separate context program loses its motivation.

\paragraph{Model scale and infrastructure value.} The bitter lesson \citep{sutton2019bitter} predicts that learned policies will replace many hand-tuned chunking rules, ordering workarounds, and fixed summarization heuristics. Proposals for \emph{agent internalization} extend this prediction to goals, identity, and deliberation \citep{xing2026critique}. Runtime acquisition and persistence serve tasks that depend on exact current repository state, sensor observations, or authorized interaction history. This yields a testable trend: as $B_\theta$ increases, representation and selection interventions should often lose value, while acquisition and persistence should retain value in proportion to a task's dependence on current external state.

\subsection{\texorpdfstring{Why $c$ Matters Now}{Why c Matters Now}}
\label{sec:why-now}

Two developments make $c$ increasingly important.

On the practical side, state created after training and private authorized state must enter through runtime infrastructure. End-to-end browsing agents combine reinforcement learning, search, tools, and evidence access \citep{openai2025deepresearch}. Their performance on GAIA \citep{mialon2023gaia} and BrowseComp \citep{openai2025browsecomp} exposes the model--infrastructure pair as the unit of performance.

On the theoretical side, Silver and Sutton \citep{silversutton2025experience} describe a shift from static human-generated data toward \emph{experience}: trajectories that agents generate through interaction with an environment. Experience updates $\theta$ as training signal and populates $c$ as grounded conditioning state at decision time. Both uses depend on infrastructure that acquires, persists, and evaluates interaction traces. The same infrastructure pattern applies to assistants, scientific tools, and knowledge-work systems outside an RL-agent setting.

\paragraph{An asymmetry with pretraining.} A web-scale corpus can support a broadly reusable base model, whereas context infrastructure is partly domain-specific: a coding repository, clinical workflow, and robot expose different acquisition surfaces, permissions, representations, and evaluators. Transferable substrate includes trace schemas, memory and retrieval primitives, access control, replay, and reporting conventions. Per-domain rebuild cost belongs in $B_c$ and shapes the economics of context investment relative to model scaling.

\paragraph{Relation to adjacent framings.}
\label{sec:adjacent-framings}

Table~\ref{tab:adjacent-framings} sets out labels in the recent literature that overlap with context scaling: what each names, its measured object, and its role in context scaling. Here, \emph{utility--cost evaluation} means measuring task utility across several $B_c$ levels under declared controls and reporting the corresponding context-infrastructure cost and any interaction with $B_d$. %Sutton's OaK proposal \citep{sutton2025oakkeynote} is one architecture for experience-driven agents, though its anti-pretraining posture differs from our complementary-axes view.

\begin{table}[htbp]
\centering
\small
\renewcommand{\arraystretch}{1.18}
\begin{tabularx}{\textwidth}{@{}>{\raggedright\arraybackslash}p{0.15\textwidth}>{\raggedright\arraybackslash}p{0.22\textwidth}>{\raggedright\arraybackslash}p{0.22\textwidth}>{\raggedright\arraybackslash}X@{}}
\toprule
\textbf{Framing} & \textbf{What it names} & \textbf{Measured object} & \textbf{Role in context scaling} \\
\midrule
Context engineering \citep{karpathy2025contextengineering,langchain2025contextengineering} & Assembly and upkeep of model-visible context & Per-call prompt / window quality & Implements context construction; context scaling measures the return on the $B_c$ spent on it \\
\addlinespace
Harness engineering \citep{he2026harnessengineering,meng2026agentharnesssurvey} & Governed runtime for action, state, recovery, observability & Runtime reliability, control, observability & Implements the runtime across the five context operations; its context costs belong to $B_c$ \\
\addlinespace
Retrieval-augmented generation (RAG) / tool use \citep{lewis2020rag,schick2023toolformer,yao2023react} & Fetching documents or tool results into $c$ & Retrieval / tool-call accuracy, end-task gain & Implements Acquire; returned evidence uses $B_c$, while query or action selection uses $B_d$ \\
\addlinespace
Memory systems \citep{memoryr1} & Persisting and recalling state across sessions & Recall accuracy over a memory horizon & Implements Persist; storage, validation, and recall are charged to $B_c$ \\
\addlinespace
Externalization\newline\citep{zhou2026externalization} & Moving memory, skills, protocols outside the weights & Where capability lives (in-weights vs.\ external) & Places state outside $\theta$; it contributes to context scaling when that state is made available through $B_c$ \\
\addlinespace
Experience-based learning \citep{suttonbarto2018rl,silversutton2025experience} & Shift toward interaction-generated learning signal & Source and yield of the learning signal & Produces traces that may update $\theta$ or populate $c$; report the resulting $B_\theta$ and $B_c$ changes separately \\
\addlinespace
Long-context windows \citep{liu2025thusspake} & Enlarging the architectural read budget & Window length $|c_{\mathrm{window}}|$ (tokens) & Raises the read limit; context scaling measures whether filling that capacity with usable state improves utility \\
\addlinespace
Agent internalization\newline\citep{xing2026critique} & Endogenous goals, identity, deliberation vs.\ external orchestration & Endogeneity of the agent's competence & Changes where competence resides, not the context budget itself; access to deployment state still uses $B_c$ \\
\midrule
\textbf{\emph{Context scaling}} & \emph{Investment in decision-time access to deployment state} & \emph{Utility--cost response to $B_c$ under declared controls} & \emph{Compares returns on these $B_c$ mechanisms while reporting interactions with $B_\theta$ and $B_d$} \\
\bottomrule
\end{tabularx}
\caption{Context scaling versus adjacent framings.}
\label{tab:adjacent-framings}
\end{table}

%Broader labels such as ``contextual AI'' or ``contextual memory intelligence'' \citep{wedel2025cmi} conflate retrieval, memory, agency, and deployment setting, and we avoid them.

%With the formal lens in place, we turn to how the system produces effective context.

\section{Context Infrastructure}
\label{sec:context-infrastructure}

Context infrastructure is the runtime system that turns deployment state into model-visible context and model decisions into observable consequences. It is the operational layer on which $B_c$ is spent. It includes stores, tool and execution interfaces, memory systems, and feedback channels that determine what the model can know, what it can do, and what the system can learn from the result.

\begin{figure}[H]
\centering
% Vector figure (default font, pdflatex build). Editable source: 14_latex_template/figures-src/fig-infra-loop.tex
\includegraphics[width=\linewidth]{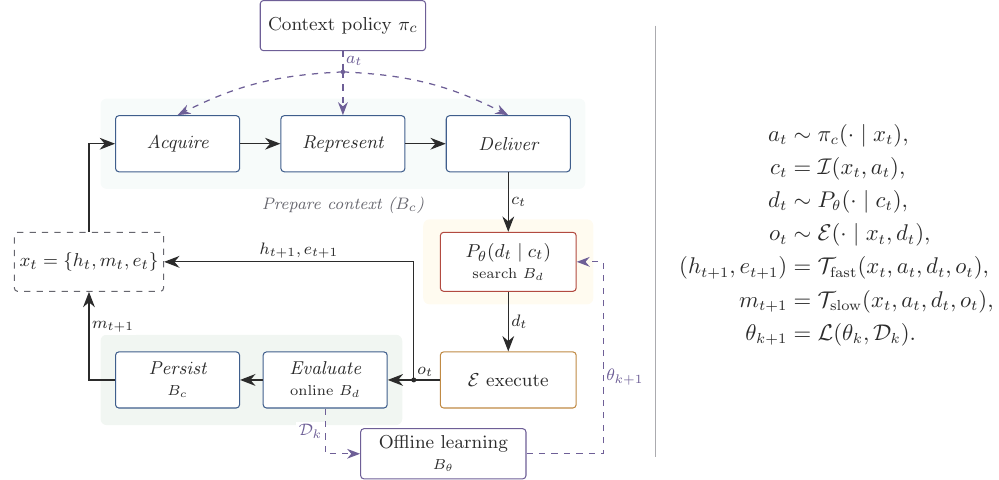}
\caption{Context infrastructure and model learning. Feedback updates fast history and environment state $(h_{t+1},e_{t+1})$, durable memory $m_{t+1}$, or model parameters $\theta_{k+1}$ from evaluated trace batches $\mathcal D_k$. Context preparation and persistence consume $B_c$, model-side search and online verification consume $B_d$, and offline learning consumes $B_\theta$; environment execution and offline evaluation costs are reported separately.}
\label{fig:context-projection}
\end{figure}

Formally, the runtime can be written as a compact partially observed interaction in which model-visible context is constructed under a budget. Let $x_t=\{h_t,m_t,e_t\}$ denote infrastructure state: history $h_t$, memory $m_t$, and environment-facing state $e_t$ (observations, tool affordances, permissions, routing). This is the step-$t$ decomposition of the deployment state $x_{\mathrm{raw}}$ above, and $\mathcal{I}(x_t,a_t)$ is the action-conditioned form of $\mathcal I_{\pi_c}$. A context policy $\pi_c$ chooses an infrastructure action $a_t$; the infrastructure produces model-visible context $c_t$; the model makes a decision $d_t$; and the execution interface returns outcome evidence $o_t$. Figure~\ref{fig:context-projection} separates the online prepare--act path from feedback that updates history and environment state, persistent memory, and model parameters.

\paragraph{Prior information-state traditions.} This framing builds on partially observable Markov decision process (POMDP) belief-state planning, active perception, and information-gathering control, which couple actions to future observations and trade task reward against information value \citep{kaelbling1998pomdp,bajcsy1988active}. Interactive information retrieval studies search as an evolving dialogue \citep{marchionini2006exploratory}, while distributed-cognition work treats external representations and artifacts as part of a cognitive system \citep{hollan2000distributed}. In LLM systems, the input available to the base model is often a serialized, permissioned, and budgeted construction assembled from histories, memories, retrievals, tool interfaces, and multimodal encoders. The context policy $\pi_c$ makes that engineering layer explicit and separately auditable from $P_\theta$.

\subsection{Context Infrastructure Across Three Update Paths}
\label{sec:amplification-loop}

Existing context-engineering practice organizes operations such as selecting, compressing, writing, and isolating context \citep{langchain2025contextengineering}. We group such operations under five lifecycle labels with distinct roles. Acquire, Represent, and Deliver form the online context-production path. Evaluate supplies measurement and credit assignment after outcomes are observed. Persist carries selected state across decisions, while evaluated traces can also feed a separate offline model-learning branch. Together they specify what state is updated, on what timescale, and under which budget.

The four failure points of \S\ref{sec:context-definition} are addressed operation by operation in the subsections below. Evaluate's online compute belongs to $B_d$ when it verifies a current decision, while its offline use belongs to evaluation or learning infrastructure and is reported separately. This accounting keeps improvements in grading or search distinct from $B_c$ interventions.

\paragraph{Content dimensions.}
\label{sec:context-dimensions}

Context content stresses the lifecycle unevenly, and four overlapping descriptive dimensions help locate the pressure: form (verbal to experiential), temporal horizon (one invocation to many sessions), modality (text to structured records, sensor, and action), and acquisition channel (observed, acted, or generated). Static text pretraining most directly covers verbal, textual records, whereas experiential and multimodal entries stress Represent, long-horizon entries stress Persist, acted channels stress Acquire and Evaluate, and generated traces stress provenance handling in Represent and Persist.

\subsubsection{Acquire: Reaching Candidate State}
\label{sec:context-acquisition}

Acquisition asks whether the relevant deployment state can be made addressable before a decision; its primary failure mode is availability failure (\S\ref{sec:context-definition}). A decisive repository file, sensor reading, tool result, user state, or environment transition remains unreachable however large the prompt. Increasing acquisition capacity broadens the candidate material from which model-visible context can be built. In agentic settings, the system may have to discover that material through a sequence of information-seeking actions under cost.

Retrieval-augmented generation and tool use bring external material into $c$ \citep{lewis2020rag,schick2023toolformer,yao2023react}. Simulators expose situated observations and transitions through interaction \citep{todorov2012mujoco}; learned world models generate approximate transitions that can be supplied as candidate context \citep{ha2018worldmodels,hafner2023dreamerv3}. When the decision procedure generates or searches learned rollouts, that compute belongs to $B_d$; selecting and packaging resulting observations or rollouts into $c$ consumes $B_c$. These channels offer different balances of controllability and fidelity; learned world-model rollouts can drift from the environment they approximate.

The most valuable sources are often privacy-sensitive and fragmented across schemas, including user histories, tool outputs, browser state, environment records, and deployment traces. Every probe also incurs latency, token, and permission costs, so wider coverage reduces the budget available to later stages of the lifecycle.

\subsubsection{Represent: Preserving Decision-Relevant Fidelity}
\label{sec:context-representation}

Acquired material must still be encoded in a form that preserves what makes it useful; failures here are fidelity failures (\S\ref{sec:context-definition}), and they occur independently of coverage. A document, video, or sensor stream may be available yet unusable because it is coarse, stale, lossy, ungrounded, or stripped of structure.

Multimodal encoders for speech, vision, video, documents, and embodied sensing determine which signals the model can absorb from raw input \citep{defossez2024moshi,liu2023llava,xu2020layoutlm,kim2024openvla}. Provenance-aware refinement, faithful summarization, and structured trace formats determine what survives the infrastructure path into $c$. A useful trace records arguments, outputs, timestamps, tool state, source authority, and permissions; a flat transcript may discard evidence needed later.

Representation loses fidelity when image downsampling removes small objects, temporal subsampling removes event order, or document linearization discards layout and relations. Text pipelines have analogous failure modes: summaries erase exceptions, refinement removes provenance, and flat transcripts lose the structure of actions and observations.

\subsubsection{Deliver: Making State Decision-Available}
\label{sec:context-delivery}

Delivery asks whether represented information reaches the model at the relevant decision point. Exclusion under the read budget is a selection failure; poor ordering, burial, or crowding can lead to an observed utilization failure. Since presentation changes the behavior of a fixed model, utilization reflects an interaction between the model and its infrastructure.

Delivery has two infrastructure decisions. The context policy determines what enters $c$, when, and in what form; a partitioning policy assigns material to context spaces. Agent harnesses rank, order, summarize, and prune material, and allocate token budgets turn by turn \citep{cursor2026selfsummarization,chroma2026context1,he2026harnessengineering,meng2026agentharnesssurvey}. Tool namespaces and sandboxes isolate sources, while sub-agents and private memories partition unrelated trajectories \citep{wu2024autogen,he2026harnessengineering,meng2026agentharnesssurvey}. Context-Folding treats branch-and-return delivery as a learned action \citep{sun2025contextfolding}. Long-context architectures and key--value (KV) cache management set the feasible read budget \citep{liu2024ropescaling,shi2024kvcachesurvey,xiao2024streamingllm}.

Access still degrades with position, length, distractors, and compression, and additional material can lower task utility \citep{liu2024lostinthemiddle,hsieh2024ruler,longppl2024}. Selection value is counterfactual, which makes it difficult to estimate before execution.

\subsubsection{Evaluate: Closing the Feedback Loop}
\label{sec:context-evaluation}

Evaluation closes the loop after execution produces evidence. Without corrective signals, acquisition accumulates noise, representation drifts, delivery admits distractors, and persistence becomes an unbounded cache. Final outcomes such as a useful answer or a passing test are coarse; signals attached to component operations can identify which acquisition, representation, delivery, or persistence decision helped \citep{lightman2023prm,cobbe2021verifier}.

Verifier, process-reward, and critique models assess intermediate steps. Infrastructure-side signals include test results, type checks, diff correctness, retrieval-faithfulness checks, user corrections, and simulator reward. These signals are most useful when they can be attributed to a specific infrastructure decision: harness-optimization systems record the expected effect of each automatic edit and compare it with later outcomes \citep{lin2026agenticharnessengineering}. Stronger verifiers permit more autonomous evaluation, while infrastructure feedback supplies data for improving the verifiers themselves.

Fine-grained credit remains expensive because it depends on labeled trajectories, simulator rewards, or detailed verifier outputs. Process rewards also invite Goodhart effects and may suppress the diversity of decisions they were intended to guide \citep{amodei2016concreteproblems,pan2022rewardmisspec}.

\subsubsection{Persist: Keeping State Across Time}
\label{sec:context-persistence}

Persistence determines which state survives beyond the current decision and how that state remains valid, recoverable, revisable, and forgettable. Missing durable state is an availability failure; stale or incorrectly summarized memory is a fidelity failure. Without persistence, the context policy must rediscover relevant state on every invocation. With it, externalized traces can shape the context and infrastructure policy used in later decisions.

External memory systems and trajectory summaries keep selected state outside the current window \citep{packer2023memgpt,zhang2024memorysurvey}. Recent work treats memory operations themselves as policy actions \citep{memoryr1,agementicmemory2026}. Staleness checks, versioning, and invalidation determine when an entry should be refreshed, superseded, or retired.

Operational data stored in persistent context may be private, biased, or product-specific. Persistent context needs authorization, provenance, audit, rollback, and forgetting hooks when state is written \citep{shavit2023agenticgovernance,casper2024audits}.

\subsection{Coding Harness as a Worked Example}
\label{sec:coding-harness-worked-example}

Production coding harnesses, including Claude Code, ChatGPT Codex, and Cursor, run all five operations in a single runtime over realistic long-horizon tasks. They are also among the few systems in which the model and surrounding infrastructure have been iterated together by the same teams; this runtime form is now formalized in engineering studies \citep{ning2026codeagentharness,pan2026naturallanguageharnesses}.

\paragraph{One runtime, five context operations.}

\begin{itemize}[leftmargin=*]
  \item \emph{Acquire} (\S\ref{sec:context-acquisition}). The harness exposes file-system reads, shell commands, repository search, version-control state, and build and test outputs, so deployment state is directly addressable.
  \item \emph{Represent} (\S\ref{sec:context-representation}). Harnesses avoid flat text dumps: they chunk files by abstract syntax tree (AST) or symbol structure and preserve tool returns with arguments, exit codes, and provenance \citep{he2026harnessengineering,meng2026agentharnesssurvey}.
  \item \emph{Deliver} (\S\ref{sec:context-delivery}). A context manager decides, turn by turn, which files, tool returns, and prior reasoning enter the next prompt under a fixed token budget; Cursor's self-summarization policy, for instance, compresses earlier turns under a task-completion reward \citep{cursor2026selfsummarization}, and sub-agent partitions keep code-review or debugging sub-tasks from contaminating the main thread.
  \item \emph{Evaluate} (\S\ref{sec:context-evaluation}). The test suite, type checker, and runtime errors provide a verifier signal that is cheap, fast, and tied to the operation that produced the artifact; user accept/reject of a diff supplies a rarer, higher-quality complement, and harness-optimization work back-attributes the combined reward to component decisions \citep{lin2026agenticharnessengineering}.
  \item \emph{Persist} (\S\ref{sec:context-persistence}). Long sessions outlast the model's context window; OpenClaw, a personal-assistant harness, uses a pre-compaction memory flush \citep{openclaw2026compaction} to write durable notes before compaction, and project-level files such as \texttt{CLAUDE.md} carry conventions across sessions while $\theta$ remains fixed.
\end{itemize}

\paragraph{Model--infrastructure co-evolution as a visible loop.} Coding is where this co-evolution is currently easiest to watch. Engineering studies and practitioner reports attribute progress on SWE-bench Verified over 2024--2026 to two axes simultaneously: stronger base models, and harness changes that alter what reaches the model in the first place \citep{jimenez2024swebench,he2026harnessengineering,lin2026agenticharnessengineering,ning2026codeagentharness}. The qualitative shape is what \S\ref{sec:context-evaluation} predicts: consented, replayable deployment traces become training signal for both the next model and the next harness.

The example is a system-level illustration of a favorable digital, replayable domain; it does not provide controlled evidence for a general response curve. Coding is also a best case along a second dimension, verification. Many economically significant deployments (creative work, mental-health support, negotiation, strategic decision-making) lack a cheap automated verifier; Evaluate becomes the bottleneck, and context interventions may be useful yet difficult to certify. Evaluation then relies on human and process signals such as interactive social-simulation outcomes \citep{zhou2024sotopia}, self-play and negotiation success \citep{fu2023negotiation}, and turn-taking and interaction quality \citep{arora2025talkingturns}, with evaluator uncertainty reported explicitly.

Beyond coding, recent AI-for-science systems run the same loop with slower evaluators and higher demands on provenance and persistence across research cycles \citep{google2026geminiscience,gottweis2026coscientist,aygun2026era}.

\subsection{The Loop as a Classifier of Mechanisms}
Table~\ref{tab:lifecycle-mechanisms} maps the representative mechanisms surveyed in this section to the operations they directly implement.

\begin{table}[htbp]
\centering
\small
\renewcommand{\arraystretch}{1.12}
\setlength{\tabcolsep}{5pt}
\begin{tabular}{@{}>{\raggedright\arraybackslash}p{0.65\textwidth}ccccc@{}}
\toprule
\textbf{Mechanism} & \textbf{Acq} & \textbf{Rep} & \textbf{Del} & \textbf{Eval} & \textbf{Per} \\
\midrule
Retrieval-augmented generation \citep{lewis2020rag}                                          & $\checkmark$ &           & $\checkmark$ &           &           \\
Tool-mediated observations \citep{yao2023react,schick2023toolformer}                         & $\checkmark$ &           &           &           &           \\
Simulator observations and selected world-model rollouts \citep{ha2018worldmodels,hafner2023dreamerv3} & $\checkmark$ &           &           &           &           \\
Multimodal encoders and document parsers \citep{liu2023llava,defossez2024moshi,xu2020layoutlm} &           & $\checkmark$ &           &           &           \\
Provenance-aware summaries and structured traces \citep{cursor2026selfsummarization,chroma2026context1} &           & $\checkmark$ & $\checkmark$ &           & $\checkmark$ \\
Context folding and branch--return policies \citep{sun2025contextfolding}                    &           &           & $\checkmark$ &           & $\checkmark$ \\
Episodic / external memory \citep{packer2023memgpt,zhang2024memorysurvey}                    &           &           & $\checkmark$ &           & $\checkmark$ \\
Memory operations as policy actions \citep{memoryr1,agementicmemory2026,mempo2026}           &           &           &           &           & $\checkmark$ \\
Process / verifier reward models \citep{lightman2023prm,cobbe2021verifier}                   &           &           &           & $\checkmark$ &           \\
Agent harness (all five operations) \citep{meng2026agentharnesssurvey,he2026harnessengineering} & $\checkmark$ & $\checkmark$ & $\checkmark$ & $\checkmark$ & $\checkmark$ \\
\bottomrule
\end{tabular}
\caption{Context-infrastructure mechanisms by lifecycle operation. Columns abbreviate Acquire (Acq), Represent (Rep), Deliver (Del), Evaluate (Eval), and Persist (Per). Checkmarks indicate direct implementation; an agent harness can implement all five operations.}
\label{tab:lifecycle-mechanisms}
\end{table}

\section{Challenges and Open Problems}
\label{sec:challenges}

The next engineering tasks fall into three clusters: \emph{grounded evaluation and reporting}, \emph{interaction data and full-cost infrastructure}, and \emph{safety and governance for persistent and adaptive systems}.

\subsection{Measurement: Evaluation Infrastructure and Proxy Metrics}
\label{sec:ch-measurement}

\paragraph{Evaluation infrastructure.} Static prompt--response benchmarks usually hold context construction outside the evaluated system. Interactive benchmarks such as SWE-bench, WebArena, and AgentBench instead require action, observation, and adaptation, though their coverage and reproducibility remain limited \citep{jimenez2024swebench,zhou2024webarena,liu2024agentbench}. Longitudinal benchmarks test memory and user state accumulated over longer horizons \citep{maharana2024locomo,zhong2024memorybank}; embodied evaluations additionally require faithful sensor and action streams.

\paragraph{Reporting.} Results should disclose the model, the infrastructure that produced $c$, separate $B_c$ and $B_d$ components, interaction and memory scale, context-quality proxies, and a utility--cost curve \citep{liu2024lostinthemiddle,hsieh2024ruler,he2026harnessengineering}. These fields should remain separate rather than being collapsed into a composite score. The checklist below defines the per-result disclosure.

\begin{tcolorbox}[colback=oursgray,colframe=black!55!white,boxrule=0.5pt,arc=2mm,left=8pt,right=8pt,top=4pt,bottom=4pt,title={\bfseries Context-scaling reporting checklist (per result)}]
\small
\begin{tabular}{@{}>{\raggedright\arraybackslash}p{0.25\textwidth}>{\raggedright\arraybackslash}p{0.65\textwidth}@{}}
\toprule
\textbf{Field} & \textbf{What to report} \\
\midrule
Model-visible context & tokens or representation units actually supplied, separate from the architectural window limit \\
$B_c$ components & storage, retrieval, tool-interface, representation, selection, persistence, latency, and maintenance costs \\
$B_d$ components & generation, search, tool-selection actions, rollouts, and online verifier compute per decision \\
Interaction depth & average actions/retrievals per committed decision \\
Memory horizon & longest span over which conditioning stays coherent \\
Decision-relevance density & optional fraction of declared units whose specified replacement lowers $U$ beyond a tolerance; report units and baseline \\
Provenance quality & source, licensing, and trust grade of admitted content \\
Recency & lag between admitted state and the current world \\
Utility--cost curve & task utility with uncertainty across several $B_c$ levels, including interactions with $B_d$ \\
\bottomrule
\end{tabular}
\end{tcolorbox}

\subsection{Interaction Data and Compute at Scale}
\label{sec:ch-data-compute}

\paragraph{Data supply.} Interaction traces grounded in environments, tools, and multimodal streams are usually produced for a specific system and tied to an environment version, unlike web text, whose original production cost was largely external to model developers. Three sources dominate. Simulation environments offer controllable volume but incur environment-authoring costs and sim-to-real gaps \citep{zhao2020sim2real,li2024simpler}. Consented product telemetry offers deployment fidelity but raises privacy and legal constraints \citep{abadi2016dpdeeplearning}. Synthetic and self-generated traces scale more readily but can narrow the data distribution as models learn from their own outputs \citep{shumailov2023curse}.

\paragraph{Full-cost infrastructure.} Closed-loop training couples policy rollouts, environment steps, evaluation, and parameter updates, and long-horizon episodes compound the cost \citep{sardana2024beyondchinchilla}. Accounting must therefore include environment simulation, evaluator queries, and trace storage alongside GPU throughput. A reusable stack needs versioned environments, structured replayable traces, privacy-preserving filters, and evaluators that can replay trajectories \citep{liu2024agentbench,zhou2024webarena,abadi2016dpdeeplearning}. Because the data is stateful and environment-dependent, the practical shared substrate is a governed collection of simulators, consented traces, and replay tools rather than a public dump.

\subsection{Safety and Governance}
\label{sec:ch-alignment-governance}

\paragraph{Safety risks.} Persistent context and adaptive policies can change deployment behavior after a model checkpoint has passed safety evaluation. Indirect prompt injection or unauthorized tool output can enter $c$ \citep{greshake2023injection}; poisoned summaries and memories can then propagate errors across sessions; weak tenant partitioning can leak private state; and repeated state or policy updates can shift refusal behavior, deference, or norm compliance \citep{perez2022redteaming,casper2024audits}. These failures can accumulate while $\theta$ remains fixed, and parameter updates add a separate source of alignment drift.

\paragraph{Governance requirements.} Governance must cover both external-state changes and model updates. A deployment should record consent and source permissions, preserve provenance for admitted and written state, version memories and model checkpoints, evaluate behavior after updates, and support rollback and forgetting \citep{shavit2023agenticgovernance,ilyas2022datamodels,c2pa2024}. Incident records should attribute harmful decisions to the retrieved passage, tool output, memory entry, policy change, or model update that preceded them. External-state rollback and model-checkpoint rollback should remain separately auditable, and material changes in behavior or data use should be disclosed to affected users and stakeholders.

\subsection{Scope and Limitations}
\label{sec:limitations}

This framework is an accounting and experimental program, not a claim that context admits a universal scalar law. The components of $B_c$ are heterogeneous and domain-dependent, so raw budget units should not be compared across task families without an explicit cost model. The evidence is also uneven: digital domains with replayable state and inexpensive verifiers are overrepresented, while embodied, interpersonal, and high-stakes settings remain harder to evaluate. Finally, the five lifecycle operations are analytical roles rather than mutually exclusive modules; a single mechanism may implement several. Establishing causal returns therefore requires controlled interventions, declared baselines, and uncertainty estimates rather than attribution from end-to-end gains alone.

\subsection{A Short Research Agenda}
\label{sec:ch-agenda}

Each direction below is empirically separable: progress on one does not wait on the others.

\begin{itemize}[leftmargin=*]
  \item \emph{Build grounded evaluation infrastructure.} The field needs reproducible, cost-controlled benchmarks for interactive and longitudinal behavior beyond static prompt--response grading.
  \item \emph{Derive principled proxies for context value.} Protocol-bound metrics such as decision-relevance density should be tested for whether they predict held-out utility better than raw context length and cost within declared task families; cross-domain comparison requires separate calibration.
  \item \emph{Develop shared interaction-data infrastructure.} Useful shared infrastructure would include versioned simulators, consented trace collection with privacy controls, synthetic-data diversity audits, replayable storage, and full-cost accounting.
  \item \emph{Establish safety and governance conventions for persistent and adaptive deployments.} Systems need tests for context-induced behavioral drift, consent and source permissions, separate audit and rollback for external state and model updates, incident attribution, and disclosure of material changes.
  \item \emph{Pursue model and context-infrastructure co-evolution.} A convincing study would estimate $U(B_\theta,B_d,B_c)$ and show how infrastructure reduces availability, fidelity, and selection failures while model improvements reduce utilization failures under matched context inputs.
\end{itemize}

\section{Conclusion}
\label{sec:conclusion}

We opened with the context bottleneck: task-relevant deployment state exists but is unavailable, distorted, poorly selected, or unused at decision time. Embedding $P_\theta(d\mid c)$ in the budgeted system distribution $Q_{\theta,\pi}(d\mid x;B_\theta,B_d,B_c)$ makes the interactions and accounting of the three budgets explicit.

Context scaling is a distinct engineering and measurement program (\S\ref{sec:ch-agenda}), evaluated through controlled budget interventions and utility--cost curves. Its engineering target is the model--infrastructure pair. The immediate work is to make $B_c$ visible in published results and to estimate $U(B_\theta,B_d,B_c)$ on declared task families; the reporting conventions, data infrastructure, and governance hooks of \S\ref{sec:challenges} are the instruments for doing so.

\FloatBarrier
\bibliographystyle{plainnat}
\bibliography{refs}

\end{document}